\documentclass[a4paper,fleqn]{cas-sc}

\usepackage[numbers,sort&compress]{natbib}

\usepackage{graphicx}
\graphicspath{{media/plots}{media/corpora}}
\makeatletter
\def\input@path{{tables/}}
\makeatother

\usepackage{amssymb}
\usepackage{amsmath}

\usepackage{multirow}
\usepackage{multicol}
\usepackage{makecell}
\usepackage{array,booktabs,arydshln}

\usepackage{subcaption}

\usepackage{hyperref}
\hypersetup{colorlinks=true,linkcolor=black}

\usepackage[acronym,nonumberlist]{glossaries}
\glsdisablehyper 

\newacronym{dl}{DL}{Deep Learning}
\newacronym{ai}{AI}{Artificial Intelligence}

\newacronym{la}{LA}{Layout Analysis}
\newacronym{al}{AL}{Active Learning}
\newacronym{omr}{OMR}{Optical Music Recognition}
\newacronym{ocr}{OCR}{Optical Character Recognition}
\newacronym{htr}{HTR}{Handwritten Text Recognition}
\newacronym{ctc}{CTC}{Connectionist temporal classification}

\newacronym{smt}{SMT}{Sheet Music Transformer}
\newglossaryentry{crnn}
{
  name={CRNN},
  description={Convolutional Recurrent Neural Network},
  first={Convolutional Recurrent Neural Network (\glsentrytext{crnn})},
  plural={CRNNs},
  descriptionplural={Convolutional Recurrent Neural Networks},
  firstplural={Convolutional Recurrent Neural Networks (\glsentryplural{crnn})}
}

\newacronym{ser}{SER}{Symbol Error Rate}

\newcommand{\Primens}{\textsc{Primens}}
\newcommand{\Capitan}{\textsc{Capitan}}
\newcommand{\Seils}{\textsc{Seils}}
\newcommand{\Guatemala}{\textsc{Guatemala}}
\newcommand{\Mottecta}{\textsc{Mottecta}}
\newcommand{\Magnificat}{\textsc{Magnificat}}

\newcommand{\GrandStaff}{\textsc{GrandStaff}}
\newcommand{\Mozarteum}{\textsc{Mozarteum}}
\newcommand{\Polonia}{\textsc{Polish Scores}}

\newcommand{\Budget}{$\mathcal{B}$}
\newcommand{\Confidence}{$\mathcal{\rho}$}

\def\tsc#1{\csdef{#1}{\textsc{\lowercase{#1}}\xspace}}
\tsc{WGM}
\tsc{QE}

\begin{document}
\let\WriteBookmarks\relax
\def\floatpagepagefraction{1}
\def\textpagefraction{.001}

\shorttitle{Sample-efficient end-to-end Optical Music Recognition through Active Learning}

\shortauthors{}  

\title [mode = title]{Sample-efficient end-to-end Optical Music Recognition through Active Learning}  



%

\author[1]{Eric Ayllon}[
      orcid=0009-0005-7468-1712
]

\cormark[1]


\ead{eric.ayllon@ua.es}


\credit{Methodology, Software, Formal analysis, Investigation, Writing - Original Draft, Writing - Review \& Editing, Visualization.}

\affiliation[1]{organization={University of Alicante},
            addressline={Carretera San Vicente del Raspeig s/n}, 
            city={San Vicente del Raspeig},
            postcode={03690}, 
            country={Spain}}

\author[1]{Francisco J. Castellanos}[orcid=0000-0001-9949-5522]


\ead{fcastellanos@dlsi.ua.es}


\credit{Methodology, Writing - Review \& Editing, Supervision, Funding acquisition.}


\author[1]{Antonio Ríos-Vila}[orcid=0000-0002-7770-9726]
\ead{arios@dlsi.ua.es}
\credit{Methodology, Writing - Review \& Editing, Supervision.}

\cortext[1]{Corresponding author}



\begin{abstract}
End-to-end approaches to Optical Music Recognition (OMR) have recently shown promising results, but their performance often degrades on unseen data. A common solution is to pretrain and fine-tune models with annotated samples from the target domain, improving adaptation to new distributions.
However, this process typically requires extensive manual annotation to yield competitive models, making it costly and time-consuming. In this work, we explore Active Learning to reduce this burden by selecting informative samples for labeling in low-resource scenarios, where the amount of manual annotation that can be performed is limited. We evaluate both uncertainty- and diversity-based sampling strategies on staff-level and full-page corpora. Results show that confidence-based sampling performs comparably to oracle selection (based on true error), and that each scenario benefits from a tailored strategy.
\end{abstract}


\begin{highlights}
\item First study of active learning techniques for Optical Music Recognition (OMR)
\item Experiments with different sampling strategies for current OMR systems
\item Confidence-based uncertainty sampling for OMR
\item Embedding-based diversity sampling for OMR
\item Results demonstrate comparable performance with limited data annotations
\end{highlights}


\maketitle

\section{Introduction}
Cultural heritage is a fundamental pillar of our society. A significant portion of musical works is physically stored in archives and libraries, where access is often restricted to prevent deterioration of the materials \citep{nikolaidou2022survey}. Digitizing and storing these documents in digital repositories addresses this issue and democratizes access. However, scanning alone is insufficient for enabling advanced analysis and interaction with musical content. Given the massive volume and considerable heterogeneity of music documents---ranging in notation style, quality, engraving, or layout, among others---automatic transcription becomes essential \citep{omr-layout-analysis,omr-with-neural-networks}, motivating the development of specialized tools for this task.

Manual transcription of historical music documents remains the most accurate method, but it is also labor-intensive and slow, becoming a major bottleneck for large-scale digitization efforts. Moreover, large datasets may still include annotation errors, especially when produced under high workloads. In this context, \gls{omr} is the research field dedicated to developing automated systems that recognize musical symbols in scanned documents and encode them into digital structured formats. Despite the progress in this field, human oversight remains essential to ensure high-quality transcriptions and minimize errors before publishing digital editions.

\gls{omr} has evolved significantly with the emergence of \gls{dl}. Traditional multi-stage pipelines \citep{multi-stage-omr} have been largely replaced by holistic neural network models \citep{omr-with-neural-networks,omr-transformer,smt}, which now represent the state of the art in \gls{omr}. By eliminating intermediate steps, these end-to-end models have notably improved transcription quality and reduced annotation efforts.
Over the years, multiple components of the \gls{omr} pipeline have been integrated into unified systems. Until recently, the predominant approach in \gls{omr} followed a two-stage process: first, a \gls{la} step extracts staff regions; then, a transcriptor individually processes their contents \citep{omr-full-page-pipeline}. This paradigm is referred to as LA-based \gls{omr}, as it explicitly identifies score regions before transcription. Recent work has proposed unifying these steps into fully holistic approaches, offering a more streamlined solution.

As \gls{ai} models increasingly take on responsibilities across the pipeline, the need for large amounts of annotated data also increases. This demand for data has become a key bottleneck, as high-quality training datasets remain scarce. Various strategies have been explored to mitigate this limitation. The most common are data augmentation and synthetic data generation. Data augmentation produces new training samples by applying transformations to existing images, simulating plausible distortions found in real-world data \citep{data-augmentation}. In contrast, synthetic data generation creates new samples by rendering musical content directly from label information, providing novel images along with ground-truth transcriptions \citep{synthetic-data}. While synthetic data provides an abundant source of training material, it often fails to capture the nuances of real-world documents. As a result, developing highly accurate music recognition systems still requires incorporating annotated real data.

Currently, developing a highly accurate music recognition system is not feasible without incorporating real-world data. However, manually annotating a sufficient number of samples to achieve competitive performance is often impractical. To mitigate the annotation burden while adapting the model to the specific characteristics of the target domain, we leverage \gls{al}---a machine learning paradigm where the most informative samples are selectively chosen for annotation, thus optimizing performance with fewer labeled examples. Following this approach, we identify and annotate promising samples from the target dataset, then fine-tune the models accordingly. Subsequently, we investigate how different AL strategies affect the performance of the fine-tuned models, with the goal of minimizing human annotation efforts by prioritizing the most informative samples \citep{deep-active-learning,active-learning-diversity}. This is especially advantageous when working with large amounts of unlabeled data \citep{active-learning-cost}, and in scenarios constrained by time, resources, or annotation budgets. Furthermore, we examine the trade-off between labeling cost and transcription quality, aiming to maximize model performance while keeping annotation demands within reasonable limits. This is particularly relevant in historical music documents, which often show high visual variability due to handwriting styles, ink degradation, or irregular layouts. These factors make it hard for models to generalize across collections. Active learning helps reduce annotation effort by selecting only the most informative samples for fine-tuning, making it well-suited to this domain.


Our experiments encompass two current \gls{omr} methodologies: LA-based and holistic. We evaluate the performance of each model on real data collections, both before and after fine-tuning them with samples selected through various \gls{al} strategies. Results show that as few as $10$ labeled samples are sufficient to produce significant improvements over the baseline in most datasets. Moreover, while \gls{al} consistently improves model performance, no single sampling strategy proves optimal across all transcription models.

The remainder of this document is organized as follows: Section~\ref{sec:background} provides background on \gls{omr} and active learning; Section~\ref{sec:methodology} introduces the problem formulation and describes the sampling strategies evaluated; Section~\ref{sec:experiments} outlines the experimental setup, including datasets, evaluation metrics, and implementation details to ensure reproducibility; Section~\ref{sec:results} presents and analyzes the results of our experiments; and finally, Section~\ref{sec:conclusions} summarizes the main findings and outlines potential directions for future research.

\section{Background}\label{sec:background}

This section reviews the foundational concepts behind our work. We begin by summarizing the evolution of \gls{omr} systems, including both traditional and modern approaches. Then, we introduce \gls{al}, a framework aimed at optimizing data selection for unseen data collections.

\subsection{Optical Music Recognition}

\gls{omr} is a research field dedicated to the digitization and transcription of musical scores. Its primary objective is to convert scanned images of sheet music into structured digital formats that faithfully represent the musical content \citep{understanding-omr,omr-review}.


Recent approaches diverge in how the recognition process is conceptualized and executed. Some focus on the explicit identification of individual symbols, followed by the analysis of their spatial and contextual relationships to derive musical interpretations, such as the work by \citet{omr-with-gnn}, which combined Faster RCNN and Graph Neural Networks to obtain a graph of elements representing the music content and their relationships, or the work by \citet{enrique_icdar2021}, whose proposal predicted the position of the next symbol from the information related to the current one with models based on VGG16 and U-Net. 

In contrast, others have embraced holistic strategies that aim to interpret the musical content directly from the raw input image. For instance, \citet{omr-with-neural-networks} introduced end-to-end neural network architectures that bypass intermediate segmentation stages, thereby reducing the annotation burden and simplifying the overall pipeline. This paradigm has been extensively explored in the literature. Representative examples include the contributions of \citet{Baro2019}, \citet{wick:2021}, and the recent work by \citet{yu2024deep}. All of these methods investigate the use of \gls{crnn} models trained with the \gls{ctc} loss, one of the most prominent and widely adopted strategies for sequence modeling over the past decade.

Due to the need for large amounts of labeled data to effectively train modern models and the limited availability of such annotated datasets in this field, a common line of research in \gls{omr} has focused on the use of synthetic data and data augmentation techniques. Synthetic datasets enable the generation of virtually unlimited training examples at minimal cost, often by rendering symbolic scores into image format using music engraving tools, which naturally provide perfectly aligned ground truth. In this sense, \citet{mayer2021} and \citet{synthetic-data} demonstrated the improvements in generalization capabilities when synthetic data is employed in this context. In addition, data augmentation has also been explored in the literature. Notable examples include the works by \citet{Calvo-Zaragoza2018b}, \citet{Calvo-Zaragoza2018}, and \citet{lopez2021data}, which show that certain types of augmentation---such as rotation or padding---are particularly beneficial for music transcription tasks. Despite the improvements in generalization brought by these techniques, they are often insufficient to produce models that are robust enough to handle real handwritten documents. In practice, a certain amount of real, annotated data is still required to adapt model weights to the specific characteristics of the target domain, especially when dealing with the variability and complexity inherent in historical or handwritten scores.

Despite these advances, modern full-page \gls{omr} pipelines still often rely on a staff retrieval stage to isolate individual staves for transcription \citep{omr-full-page-pipeline}. Motivated by the advances in Handwritten Text Recognition (HTR), an analog field to \gls{omr} in the textual context, recent work has explored segmentation-free models for \gls{omr}. \citet{smt,smtpp} proposed the first full-page, segmentation-free architectures capable of transcribing polyphonic scores. These models represent a promising step forward by reducing the dependency on layout analysis and minimizing the need for annotated regions. Their main obstacle is the need for a massive amount of labeled data, which has led to the consideration of pre-training on synthetic data before performing specialization adjustments to new domains, with the corresponding annotation.

\subsection{Active learning}

The performance of automatic transcription systems heavily depends on the data used to fine-tune their parameters. While it is possible to improve generalization to unseen collections by labeling a subset of samples and fine-tuning the model on them, recent trends show that the performance of modern architectures tends to increase with the amount of training data available \citep{data-augmentation}. However, to truly leverage large pools of data, each selected sample must be relevant and contribute new information. Therefore, an informed selection process becomes essential.

\gls{al} focuses on the development of strategies for algorithmically identifying which samples should be selected for annotation and training \citep{deep-active-learning,pool-based-active-learning}. Once selected, these samples are labeled by a human oracle. The overarching goal of \gls{al} is to maximize learning efficiency while minimizing annotation effort \citep{active-learning-cost,transformer-active-learning}. A common setup involves defining a fixed limit on the number of samples that can be annotated from a large collection of unlabeled data, typically referred to as pool-based \citep{active-learning-budget-2022,active-learning-budget-2023}.

Most prominent AL methods in the literature fall into two main categories: uncertainty-based sampling and diversity-based sampling. Uncertainty sampling prioritizes samples for which a probabilistic model exhibits low confidence in its predictions. This approach is grounded in two main hypotheses: first, that highly uncertain samples are inherently more informative \citep{uncertainty-sampling-1,uncertainty-sampling-2}; and second, that informativeness alone is sufficient for building an effective training set. However, a known limitation of uncertainty-based strategies is their tendency to select similar samples repeatedly, leading to redundancy in the training data \citep{hybrid-sampling}. Diversity sampling, on the other hand, aims to select dissimilar samples to increase the variety of training instances. This strategy assumes that similar samples convey redundant information and are less likely to cover edge cases within the distribution \citep{active-learning-finetuning}. Nonetheless, this approach may include outlier or uninformative samples that do not contribute meaningfully to model improvement \citep{hybrid-sampling}. To address the limitations of single-criterion methods, recent research has explored hybrid strategies that combine uncertainty and diversity. Some approaches use a weighted sum of both criteria, while others introduce adaptive mechanisms. For example, \citet{adaptive-hybrid} proposed a self-adjusting scheme to balance uncertainty and diversity dynamically, whereas \citet{hybrid-sampling} relied on a fixed hyperparameter to combine them. Although these hybrid methods introduce greater complexity, single-criterion strategies have often shown more consistent performance in practice, as described in the survey by \citet{deep-active-learning}.

\section{Methodology}\label{sec:methodology}
This section outlines the methodology employed in our experiments. We begin by introducing a general formulation of the transcription process, which serves as a common framework for both \gls{la}-based and \gls{la}-free pipelines. We then describe how the pool-based \gls{al} paradigm---widely used in this domain---is formalized within our setting. 

It is important to note that our focus lies exclusively on the transcription stage. In the \gls{la}-based approach, as is customary in related work, we assume that layout analysis has already been performed, and the transcription model operates on isolated single-staff images. In contrast, the \gls{la}-free approach processes full-page images that may contain multiple staves or systems. 

\subsection{General formulation of music transcription}
The typical input to an \gls{omr} system is an image of a musical score or staff, and the expected output is a sequence of symbols that encodes its visual content. To formalize this process, we define an image space $\mathcal{X}$ and a vocabulary $\Sigma$ comprising all possible symbols. The output space is thus the set of all finite-length sequences over this vocabulary, denoted $\Sigma^*$. For simplicity, we assume that any special tokens (e.g., padding or sequence delimiters) are already included in $\Sigma$; such symbols will be explicitly mentioned when relevant. The transcription task can therefore be framed as learning a mapping $f: \mathcal{X} \rightarrow \Sigma^*$.

Given that both data availability and computational resources are limited in practice, the function $f$ must be approximated through supervised learning. The specific form of this approximation depends on the pipeline being used. In the following sections, we detail the models and training procedures used in both the \gls{la}-based and \gls{la}-free settings.

\subsubsection{Staff-level transcription}

Current state-of-the-art \gls{omr} systems typically require a preliminary layout analysis step to extract image regions containing individual staves or systems \citep{omr-full-page-pipeline}. In this pipeline, we assume that the input to our transcription model consists of these isolated staff images. Working with such localized regions enables certain assumptions and simplifications. For instance, in the case of monophonic music, there is only one active voice at a time, implying that no concurrent symbols appear in the image. As a result, the transcription process can be formulated similarly to line-level handwritten text recognition, which typically employs a \gls{crnn} architecture trained with the \gls{ctc} loss function to approximate the transcription function $f$ \citep{htr-survey}. This architecture consists of a convolutional encoder followed by a recurrent decoder. The encoder receives a normalized input image $x \in [0,1]^{h \times w \times c}$ and extracts visual features, producing a latent representation. This representation is then passed to the decoder, which outputs a sequence of symbol probability distributions:

\begin{equation}\label{eq:posteriorgram}
    P(\sigma \mid e_{\text{crnn}}(x)_t), \quad 1 \le t \le w',\; \sigma \in \Sigma,
\end{equation}
where $e_{\text{crnn}}(x)_t$ denotes the $t$-th frame in the encoded sequence.

A softmax activation is applied at each time step to yield a posterior distribution over the vocabulary $\Sigma$. The resulting sequence of distributions is commonly referred to as the \emph{posteriorgram} of $x$. The \gls{ctc} loss enables the model to learn an alignment between input and output sequences without requiring frame-level annotations. A special blank symbol $\epsilon$ is added to $\Sigma$ to allow the model to handle temporal stretching and to differentiate repeated symbols.

During inference, the model outputs a sequence of predicted symbols by selecting the most probable symbol at each time step via greedy decoding:

\begin{equation}
    \hat{\sigma}_t = \arg\max_{\sigma \in \Sigma} P(\sigma \mid e_{\text{crnn}}(x)_t),
\end{equation}
yielding the raw prediction sequence $\hat{\sigma}_1, \hat{\sigma}_2, \ldots, \hat{\sigma}_{w'}$. Post-processing involves collapsing consecutive repeated symbols and removing all occurrences of the blank symbol $\epsilon$, resulting in the final transcription hypothesis \citep{ctc}.

\subsubsection{Full-page OMR}

Despite the high performance of the \gls{crnn} approach, the staff extraction step can become an additional source of errors, often requiring human supervision to ensure that all regions of interest are correctly detected and extracted. Moreover, \gls{la}-based methods face limitations when transcribing complex scores such as polyphonic music \citep{polyphonic-scores} or pieces containing lyrics \citep{OMR-with-lyrics}. These cases typically require a post-transcription alignment step, which can be challenging \citep{alignment} and substantially increase the effort needed to correct potential errors in the final output.

In contrast to the staff-level pipeline, the \gls{smt} architecture \citep{smt} enables end-to-end transcription directly from full-page images, without requiring prior staff detection \citep{smtpp}. This model consists of a convolutional encoder---specifically, a ConvNeXt---and a transformer decoder capable of handling simultaneous aligned sequences.

The visual feature extraction process is similar to that of the \gls{crnn} encoder. However, the \gls{smt} arranges the latent representation differently, producing a mapping of the form $e_\text{smt}: \mathcal{X} \rightarrow \mathbb{R}^{h' \cdot w' \times c'}$. Before flattening the spatial dimensions ($h'$ and $w'$), a two-dimensional positional encoding~\citep{document-attention-network} is added to the feature map to provide spatial context to the transformer decoder.

During training, the decoder receives both the encoded image and the ground-truth token sequence. At inference time, however, the transcription must be generated autoregressively, since the target sequence length is unknown. A special start-of-sequence token $\hat{\sigma}_0$ is prepended to the input, and a reserved end-of-sequence token is included to signal termination. In cases where the decoder fails to generate this token, a maximum number of decoding steps is enforced to prevent infinite loops.

The output of the \gls{smt} model can be formally described as:

\begin{equation}
    P(\sigma\mid e_\text{smt}(x), (\hat{\sigma}_0,\dots,\hat{\sigma}_{t-1})) \quad \sigma \in \Sigma,
\end{equation}
where $\hat{\sigma}_t$ denotes the predicted symbol at decoding step $t$. A greedy decoding strategy is employed, selecting the most probable token at each step:

\begin{equation}
    \hat{\sigma}_t = \arg\max_{\sigma \in \Sigma} P(\sigma\mid e_\text{smt}(x), (\hat{\sigma}_0,\dots,\hat{\sigma}_{t-1})).
\end{equation}

Once the start and end tokens are removed, the decoder output corresponds to a valid transcription sequence.

\subsection{Active Learning}

Pool-based active learning defines a process that starts with two disjoint sets: one containing labeled data, and one containing unlabeled data. Samples from the unlabeled set are algorithmically selected for annotation in successive rounds. These sets are typically denoted as $\mathcal{D}_l = \left\{(x_1, y_1), \dots, (x_n, y_n)\right\}$ for labeled data, and $\mathcal{D}_u = \left\{x_1, \dots, x_m\right\}$ for unlabeled data.

A batch $\mathcal{D}_q \subset \mathcal{D}_u$ is selected for annotation by an oracle and subsequently moved from $\mathcal{D}_u$ to $\mathcal{D}_l$. If we define the oracle labeling function as $l: \mathcal{X} \rightarrow \Sigma^*$, which returns the true label of a given sample $x \in \mathcal{X}$, the update can be formally described as in Equation~\ref{eq:dataset}
\begin{equation}\label{eq:dataset}
\mathcal{D}_l' = \mathcal{D}_l \cup \{(x, l(x)) : x \in \mathcal{D}_q\}, \quad \mathcal{D}_u' = \mathcal{D}_u \setminus \mathcal{D}_q.
\end{equation}
The model is then retrained using the updated labeled set $\mathcal{D}_l'$. The total number of queries allowed is bounded by a labeling budget \Budget{} \citep{pool-based-active-learning,active-learning-finetuning}.

Our methodology is inspired by the work of \citet{active-learning-finetuning}, where training begins with $\mathcal{D}_l = \emptyset$ and relies on a pre-trained model that is fine-tuned using the progressively augmented labeled set $\mathcal{D}_l'$. To ensure adaptability across different datasets, we define the output vocabulary $\Sigma$ as the union of the vocabulary from the pre-training dataset and that of the labeled examples in $\mathcal{D}_l'$.

\subsection{Uncertainty estimation}

A key challenge in active learning is determining which unlabeled samples are most informative for the model. Since annotating data is costly, randomly selecting instances from $\mathcal{D}_u$ may lead to suboptimal use of the labeling budget. To address this, various selection strategies have been proposed that aim to prioritize samples the model is uncertain about. These methods require a reliable estimation of model uncertainty, which can then be used as a criterion to rank or filter candidates for annotation. In this work, we adopt strategies based on the predictions due to their simplicity and proven effectiveness \citep{active-learning-difficulty-confidence-based}. 

The confidence of an individual predicted token is defined as the posterior probability assigned to the predicted class \citep{uncertainty-sampling-1}. To extend this definition to sequence labeling tasks, we follow the approach proposed by \citet{htr-omr-calibration}, which approximates the confidence of an entire sequence by averaging the confidence scores of the individual tokens. 

This approximation allows us to define the confidence \Confidence{} of a prediction in \gls{omr} as follows. For the \gls{crnn} model:

\begin{equation}
    \rho_\text{crnn}(x) = \frac{1}{w'}\sum_{t=1}^{w'} p(\hat{\sigma}_t \mid e_\text{crnn}(x)_t)~,
\end{equation}
where $w'$ denotes the temporal width of the model output, i.e., the number of frame-level predictions. For the \gls{smt} model, confidence is defined as:

\begin{equation}
    \rho_\text{smt}(x) = \frac{1}{\vert\hat{\sigma}\vert}\sum_{t=1}^{\vert\hat{\sigma}\vert}p(\hat{\sigma}_t \mid e_\text{smt}(x), (\hat{\sigma}_0,\dots,\hat{\sigma}_{t-1}))~,
\end{equation}
where $\hat{\sigma}$ denotes the predicted sequence and $\vert\hat{\sigma}\vert$ its length.

\subsection{Sampling strategies}

The selection of samples from the unlabeled set $\mathcal{D}_u$ plays a crucial role in the effectiveness of active learning. Since annotation resources are limited, choosing the most informative examples can significantly impact model performance with the minimum annotation effort. In this work, we examine several selection criteria, including uncertainty-based and diversity-based methods. The following sections describe how these strategies are computed and applied within our framework.

\subsubsection{Uncertainty sampling}\label{sec:uncertainty-sampling}

Many uncertainty-based active learning methods focus on selecting the most uncertain samples \citep{active-learning-uncertainty,uncertainty-sampling-1,active-learning-difficulty-based,active-learning-difficulty-confidence-based}. While effective, this approach can lead to the selection of outliers if the labeled set $\mathcal{D}_l$ is not representative. To mitigate this, we also explore sampling at intermediate uncertainty levels, such as near the mean or specific quartiles, which may offer a better balance between informativeness and reliability. These strategies aim to avoid noisy or redundant samples while still improving model performance \citep{active-learning-difficulty-confidence-based}.

To address this, we explore multiple uncertainty-based sampling criteria. Each strategy is formalized as a score function $s(x)$ that assigns a value to each sample $x \in \mathcal{D}_u$. The \Budget{} samples with the lowest scores are selected for annotation. For clarity, we denote the predicted sequence for input $x$ as $\hat{\sigma}^x$, and define $\rho(x)$ as the prediction confidence, regardless of whether the underlying model is a \gls{crnn} or a \gls{smt}.

\begin{itemize}
    \item \textbf{Highest uncertainty}. Selects the samples with the lowest confidence, based on the assumption that lower confidence correlates with higher prediction error.
    \begin{equation}
        s(x) = \rho(x) \qquad s(x) = -\text{SER}(\hat{\sigma}^x, l(x))
    \end{equation}

    \item \textbf{Near} $\mathbf{Q_3}$. Selects samples whose uncertainty is closest to the third quartile of the uncertainty distribution over $\mathcal{D}_u$.
    \begin{equation}
        s(x) = \left|1 - \rho(x) - Q_3^{1-\rho} \right| \qquad 
        s(x) = \left| \text{SER}(\hat{\sigma}^x, l(x)) - Q_3^\text{SER} \right|
    \end{equation}

    \item \textbf{Near mean}. Selects samples whose uncertainty is closest to the mean uncertainty in $\mathcal{D}_u$.
    \begin{equation}
        \begin{array}{c}
            s(x) = \left| \rho(x) - \frac{1}{|\mathcal{D}_u|} \sum_{x' \in \mathcal{D}_u} \rho(x') \right| \\ \\
            s(x) = \left| \text{SER}(\hat{\sigma}^x, l(x)) - \frac{1}{|\mathcal{D}_u|} \sum_{x' \in \mathcal{D}_u} \text{SER}(\hat{\sigma}^{x'}, l(x')) \right|
        \end{array}
    \end{equation}

    \item \textbf{Near} $\mathbf{Q_1}$. Selects samples closest to the first quartile of uncertainty in $\mathcal{D}_u$.
    \begin{equation}
        s(x) = \left|1 - \rho(x) - Q_1^{1-\rho} \right| \qquad 
        s(x) = \left| \text{SER}(\hat{\sigma}^x, l(x)) - Q_1^\text{SER} \right|
    \end{equation}

    \item \textbf{Lowest uncertainty}. Selects the samples the model is most confident about.
    \begin{equation}
        s(x) = 1 - \rho(x) \qquad s(x) = \text{SER}(\hat{\sigma}^x, l(x))
    \end{equation}
\end{itemize}

The sampling function is then defined as:
\begin{equation}
    \text{Sample}(\mathcal{D}_u, B) = \left\{ \mathcal{D}_u^i : 1 \le i \le B \right\},
\end{equation}
such that
\begin{equation}
    i < j \rightarrow s(\mathcal{D}_u^i) \le s(\mathcal{D}_u^j), \quad 1 \le i, j \le |\mathcal{D}_u|.
\end{equation}

That is, all samples in $\mathcal{D}_u$ are sorted in ascending order according to $s(\cdot)$, and the \Budget{} lowest-scoring ones are returned.

\subsubsection{Diversity sampling}

For diversity-based sampling, we adopt the same strategy proposed by \citet{active-learning-finetuning}, which aims to construct the query set $\mathcal{D}_q$ such that its elements are maximally dissimilar from each other, while still being representative of the overall distribution in $\mathcal{D}_u'$.

Let $\phi(x)$ denote a feature representation extracted from each sample $x \in \mathcal{D}_u$. The goal is to find a set of \Budget{} centroids $\hat{\Theta} = \{\hat{\theta}^1,\dots,\hat{\theta}^{\mathcal{B}}\}$ that maximize the similarity between each sample and its closest centroid:

\begin{equation}\label{eq:similarity}
    \hat{\Theta} = \arg\max_\Theta S(\phi(x), \dot{\theta}_x) = \mathop{E}_{x \in \mathcal{D}_u} \left[ \frac{\text{sim}(\phi(x), \dot{\theta}_x)}{\tau} \right]~,
\end{equation}
where $\tau$ is a temperature parameter, $\text{sim}(\cdot, \cdot)$ denotes cosine similarity, and $\dot{\theta}_x = \arg\max_{\theta \in \Theta} \text{sim}(\phi(x), \theta)$ is the closest centroid to $\phi(x)$.

However, this objective may lead to a ``collapse'', where all centroids converge to the same point, i.e., $\theta^i = \theta^j$ for all $i, j$. To prevent this, a regularization term is introduced to penalize similarity between centroids and thus promote \textit{diversity} among them:

\begin{equation}
    \text{Diversity}(\Theta) = -\mathop{E}_{1 \le i \le \mathcal{B}} \left[ \log \sum_{\substack{j=1 \\ j \ne i}}^{\mathcal{B}} \exp\left(\frac{\text{sim}(\theta^i, \theta^j)}{\tau}\right) \right].
\end{equation}

By adding this term, the optimization objective becomes:

\begin{equation}
    \hat\Theta = \arg\max_\Theta \left[ S(\phi(x), \dot\theta_x) + \text{Diversity}(\Theta) \right].
\end{equation}

Once the centroids $\hat{\Theta}$ are obtained, the sampling function selects the samples in $\mathcal{D}_u$ that are closest to each centroid:
\begin{equation}
    \text{Sample}(\mathcal{D}_u) = \left\{ \arg\max_{x \in \mathcal{D}_u} \text{sim}(\phi(x), \theta) : \forall \theta \in \hat\Theta \right\}.
\end{equation}

In our framework, we use the encoder as the feature extractor $\phi$, providing a latent representation based on the visual content of the input image.

\section{Experiments}\label{sec:experiments}

Our experimental pipeline is divided into three steps. First, we pre-train an \gls{omr} model using a synthetic dataset to obtain a baseline. Next, we select the budget of \Budget{} images $\mathcal{D}_q$ from a target unlabeled dataset $\mathcal{D}_u$ using one of the presented selection criteria, obtain their true labels, and add them to $\mathcal{D}_l$, which is initially empty. Finally, we fine-tune the transcription model using the updated labeled set $\mathcal{D}_l'$.

The following sections provide implementation details to ensure reproducibility.\footnote{The source code will be made publicly available upon publication}

\subsection{Corpora}

To pre-train the models, we used \Primens{} \citep{primens} and \GrandStaff{} \citep{smtpp}. These collections are not used for evaluation. \Primens{} consists of 4,923 images of monophonic staves in mensural notation, compiled from works by Agricola, Frye, and Ockeghem, and available through the Josquin Research Project.\footnote{\url{https://josquin.stanford.edu/}} \GrandStaff{} contains 688 full-page images of piano-form music by six well-known composers, available in the Humdrum repository.\footnote{\url{https://github.com/humdrum-tools/humdrum-data}} Images from both collections were rendered using the Verovio digital engraver \citep{verovio}.

\begin{figure}[!htbp]
    \centering
    \begin{subfigure}{.42\textwidth}
        \includegraphics[width=.9\textwidth]{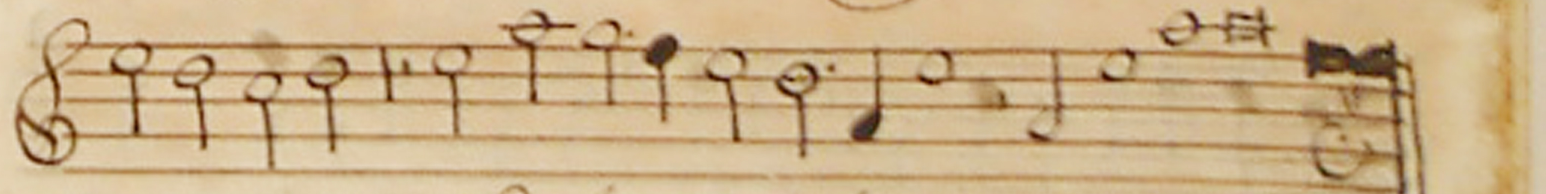}
        \includegraphics[width=.9\textwidth]{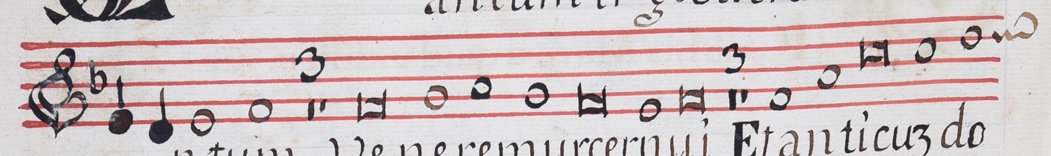}
        \includegraphics[width=.9\textwidth]{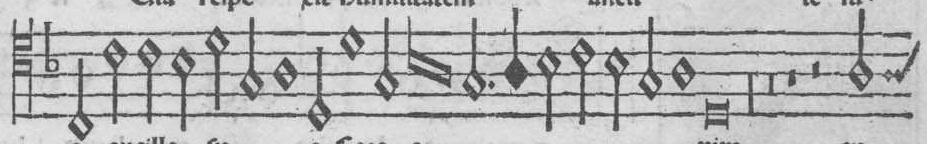}
        \includegraphics[width=.9\textwidth]{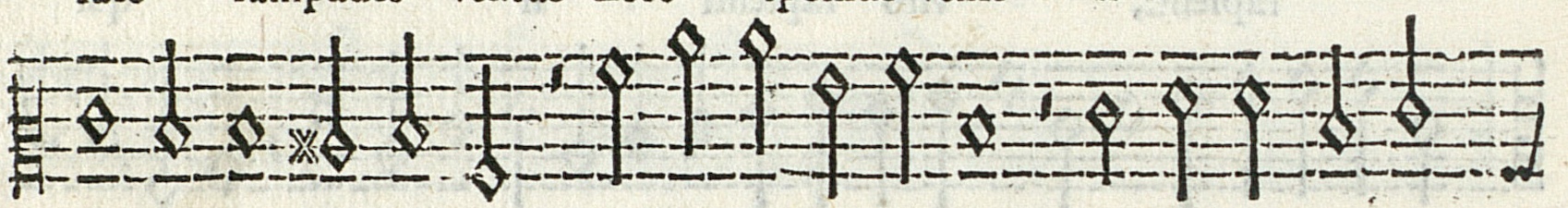}
        \includegraphics[width=.9\textwidth]{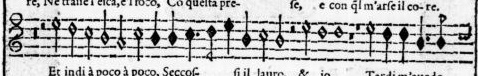}
    \end{subfigure}
    \begin{subfigure}{.23\textwidth}
    \includegraphics[width=.9\textwidth]{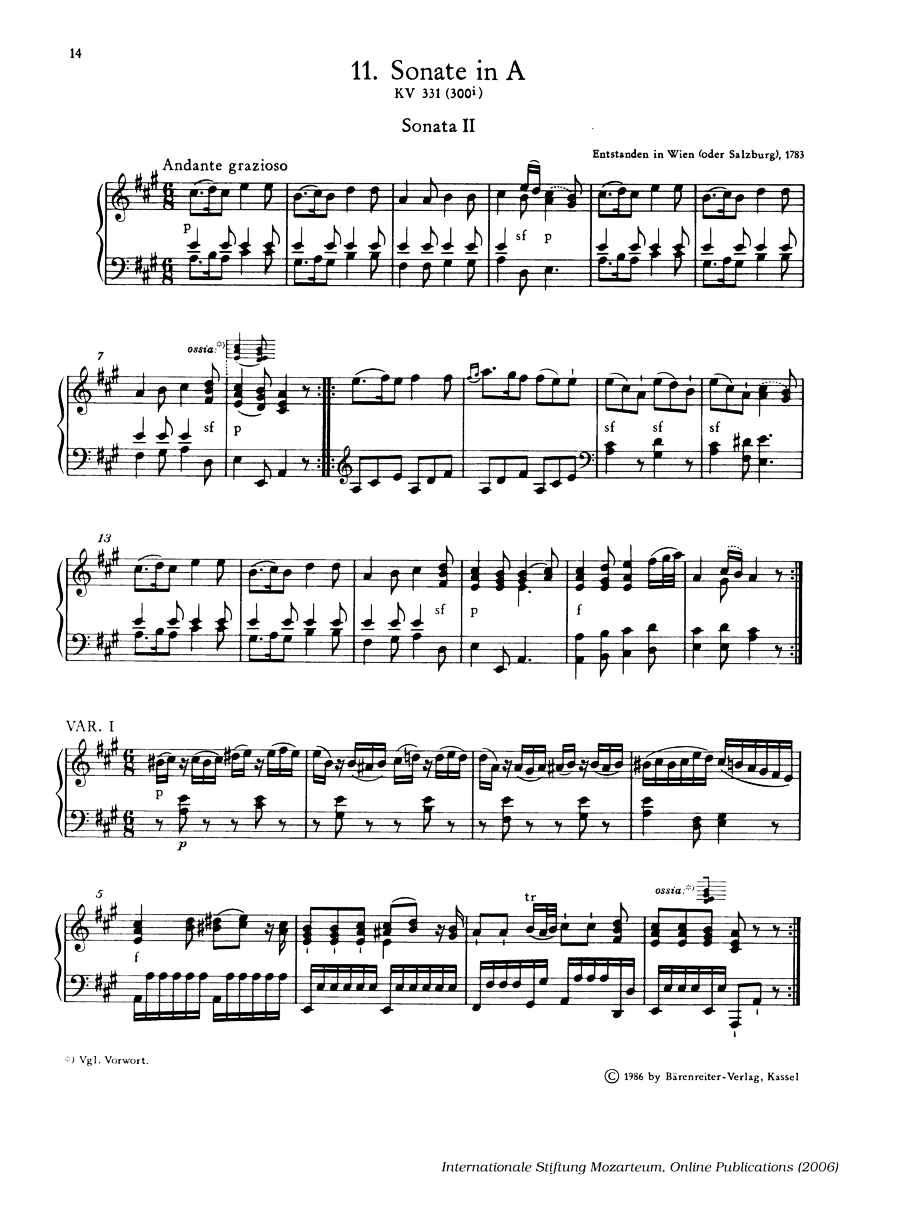}
    \end{subfigure}
    \begin{subfigure}{.23\textwidth}
    \includegraphics[width=.9\textwidth]{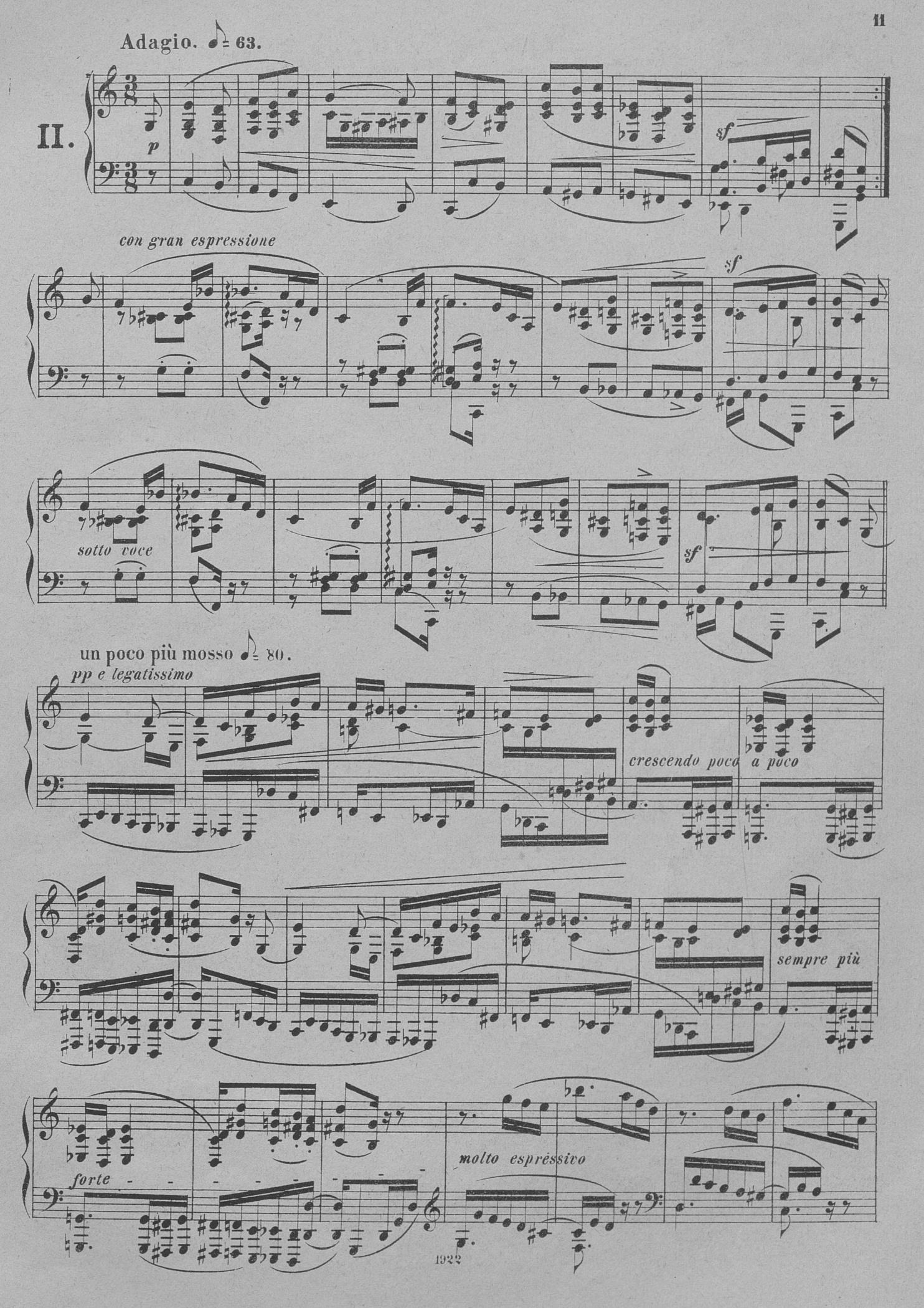}
    \end{subfigure}
    \caption{Example images from the staff-level collections (left), \Mozarteum{} (middle) and \Polonia{} (right). The images from the staff-level collections are from top to bottom: \Capitan{}, \Guatemala{}, \Magnificat{}, \Mottecta{}, and \Seils{}.}
    \label{fig:corpora-examples}
\end{figure}

For model evaluation, we used five staff-level datasets and two full-page collections, described below. Table~\ref{tab:corpora} summarizes key properties of each dataset, and Figure~\ref{fig:corpora-examples} provides an example image of each collection to better understand the variety of collections considered in this study.

Concerning the staff-level collections, we considered the following list:
\begin{itemize}
    \item \Capitan{} \citep{capitan}: Monophonic scores from a 17th-century \textit{missa} in mensural notation.
    \item \Guatemala{} \citep{guatemala}: A subset of mensural scores from six Guatemalan choirbooks (17th–18th centuries), from the Cathedral of Guatemala.
    \item \Magnificat{} \citep{unsupervised-domain-adaptation}: Monophonic mensural scores from a 16th-century Spanish choir book.
    \item \Mottecta{} \citep{magnificat-mottecta}: Scores from Francisco Guerrero's \textit{Mottecta}, originally written in the 16th century and edited by Giacomo Vicenti in the 17th century. Publicly available via the Biblioteca Digital Hispánica.\footnote{\url{https://bdh.bne.es/bnesearch/detalle/bdh0000008932}}
    \item \Seils{} \citep{seils}: Monophonic scores from the \textit{Il lauro Secco} collection, a 16th-century Italian madrigal.
\end{itemize}

For the full-page scenario, we used the following collections:

\begin{itemize}
    \item \Mozarteum{}: Pianoform scores in modern notation from the \textit{Digital Mozarteum Edition}, which provides a publicly available re-edition of works by Wolfgang Amadeus Mozart. The scores are accessible via the MoVi tool.\footnote{\url{https://dme.mozarteum.at/movi/en}}
    
    \item \Polonia{} \citep{polish-scores}: A subset of 117 pianoform pages from the \textit{Polish Digital Scores}, a repository of Polish music compiled as part of the \textit{Heritage of Polish Music in Open Access} project led by the Chopin Institute.\footnote{\url{https://github.com/pl-wnifc/humdrum-polish-scores}}
\end{itemize}

\begin{table}[hbt]
    \centering
    \renewcommand{\arraystretch}{1}
    \setlength{\tabcolsep}{5pt}
    \begin{tabular}{ll c cc c ccccc}
    \toprule[1pt]
    \multicolumn{2}{l}{\multirow{2}{*}{\textbf{Dataset}}} &  & \multicolumn{2}{c}{\textbf{Symbols}} &\phantom{---}& \multicolumn{3}{c}{\textbf{Sequence length}} & \multirow{2}{*}{\makecell{\textbf{\# of} \\ \textbf{samples}}} & \multirow{2}{*}{\makecell{\textbf{Handwritten} \\ \textbf{or Printed}}}\\
    \cmidrule{4-5} \cmidrule{7-9}
     &&& \textbf{Unique} & \textbf{Total} && \textbf{Min} & \textbf{Average} & \textbf{Max} && \\
    \cmidrule{1-11}
    \multicolumn{11}{l}{\textbf{\textit{Staff-level}}} \\
    & \Capitan{} && 323 & 17112 && 8 & 24.3 & 35 & 704 & H \\
    & \Guatemala{} && 252 & 78558 && 4 & 24.1 & 49 & 3263 & H \\
    & \Magnificat{} && 181 & 40085 && 2 & 29.5 & 53 & 1361 & P \\
    & \Mottecta{} && 199 & 43192 && 5 & 23.4 & 41 & 1845 & P \\
    & \Seils{} && 185 & 30689 && 5 & 28.8 & 48 & 1067 & P \\
    \cmidrule{1-11}
    \multicolumn{11}{l}{\textbf{\textit{Full-page}}} \\
    & \Mozarteum{} && 191 & 122477 && 32 & 2006 & 3264 & 101 & P \\
    & \Polonia{} && 215 & 141866 && 1103 & 2196 & 4116 & 117 & P \\
    \bottomrule[1pt]
    \end{tabular}
    \caption{Overview of the corpora used in the experiments, including relevant properties of their samples, such as the number of symbols, the average sequence length (or number of tokens) of the ground truth sequences, total number of samples, and whether their scores are handwritten (H) or printed (P).}
    \label{tab:corpora}
\end{table}

For the experiments, we reserved 10\% of the samples from each collection as a held-out test set. No sample from any test set was allowed to be selected for fine-tuning. These samples were used exclusively to evaluate model performance, both before and after fine-tuning.

\subsection{Metrics}\label{sec:metrics}

Following the standard in \gls{omr} \citep{omr-review}, and considering that the corpora considered contain agnostic encoding, we evaluate the performance of our system using the \gls{ser}. This metric is defined as the edit distance \citep{levenshtein-distance} between the predicted and ground-truth sequences, normalized by the length of the ground-truth sequence. This metric reflects the number of edits required to transform the predicted sequence into the correct one, thus quantifying the effort needed for manual correction. Let $\hat{\sigma}^x$ denote the predicted sequence for a given image $x$, and $l(x)$ its ground truth. Then, the \gls{ser} is defined as in Equation~\ref{eq:ser}.

\begin{equation}\label{eq:ser}
    \text{SER}(\hat{\sigma}^x, l(x)) = \frac{\text{ED}(\hat{\sigma}^x, l(x))}{\vert l(x) \vert}
\end{equation}
where ED represents the edit distance between two sequences.

To facilitate the analysis of budget efficiency, we introduce an auxiliary metric called \textit{Improvement Progress}, which quantifies the reduction in \gls{ser} achieved at each budget level, relative to two reference points: the pre-trained model, which serves as the initial baseline and whose performance corresponds to the \textit{zero-shot} setting (i.e., without fine-tuning); and the oracle model, obtained by fine-tuning the pre-trained model using the entire training set from the initially unlabeled collection. 

The metric is defined as the ratio between the error reduction achieved at a given budget and the maximum possible error reduction (i.e., that achieved by the oracle model), both normalized with respect to the zero-shot performance (see Equation~\ref{eq:progress}).

\begin{equation}\label{eq:progress}
\text{Progress} = \frac{1-\text{SER} / \text{SER}_\text{zero-shot}}{1-\text{SER}_\text{ideal-model} / \text{SER}_\text{zero-shot}}
\end{equation}

\subsection{Implementation details}\label{sec:implementation-details}

The \gls{crnn} model consists of a convolutional encoder with four blocks, each containing 64 filters, followed by a recurrent decoder composed of two LSTM layers and a final linear projection, following the architecture proposed by \citet{htr-omr-calibration}. 

For the \gls{la}-free scenario, we use the \gls{smt} model with the same configuration as in \citet{smtpp}. The encoder is a ConvNeXt network with three convolutional blocks of hidden sizes 64, 128, and 256, and depths 3, 3, and 9, respectively. The decoder is a transformer with eight layers and a hidden size of 256.

The \gls{crnn} is used for staff-level transcription, while the \gls{smt} handles full-page input. Since the goal of this study is to evaluate the effect of active learning on fine-tuning, no task-specific hyperparameter tuning was performed. To minimize variability and ensure a fair comparison between selection criteria, the weights of the pre-trained models are fixed during the sample selection step and reused as the initialization point for fine-tuning in all experiments. To control stochastic effects such as data augmentation and dropout during fine-tuning, the random seed is consistently set to 42 using PyTorch Lightning’s \texttt{seed\_everything} function. Importantly, these sources of randomness only affect the fine-tuning stage, as the sample selection process is fully deterministic.

We evaluate multiple annotation budgets to explore the trade-off between labeling cost and performance gain. Specifically, we consider $\mathcal{B} \in \{10, 25, 50\}$ for all experiments, and additionally $\mathcal{B} = 100$ in the staff-level scenario only, since the training and validation sets in \Mozarteum{} and \Polonia{} contain fewer than 100 samples.

To increase the diversity of the fine-tuning data, we apply data augmentation to the selected samples $\mathcal{D}_q$. For the \gls{crnn}, we randomly apply affine transformations with a 50\% probability, including rotations between $-3^\circ$ and $3^\circ$, and translations up to 5\% of the input’s height and width.\footnote{See documentation: \url{https://pytorch.org/vision/main/generated/torchvision.transforms.v2.RandomAffine.html}} For the \gls{smt}, we use the same augmentations described in the work by \citet{smtpp}, whose implementation is publicly available.\footnote{\url{https://github.com/antoniorv6/SMT-plusplus}}

After selecting $\mathcal{D}_q$, the remaining samples in $\mathcal{D}_u'$ (excluding those reserved for testing) are used as a validation set, providing an approximate lower bound of the model’s \gls{ser}.

While \gls{al} approaches may involve multiple rounds of sampling and labeling, our experimental setup follows the one-shot strategy proposed by \citet{active-learning-finetuning}, performing a single iteration of selection and fine-tuning with exhaustive experimentation in that iteration. We chose this approach for three main reasons: First, we consider it unfair to compare single-iteration diversity sampling with multi-iteration uncertainty sampling, as this would introduce methodological imbalance; second, limiting the process to a single iteration minimizes time, computational, and annotation costs, as expert labeling is only required after a single sampling step; and third, adding multiple iterations allows stochastic artifacts from the fine-tuning step to influence the sampling protocol, which would lose its determinism. By keeping the sampling process as deterministic as possible, we expect the analysis and conclusions to be more reliable, allowing us to extrapolate them to practical applications. Furthermore, in multi-iteration Active Learning experimental setups, the number of iterations also affects the final performance; thus, a detailed study of this variable would be required. Thus, to prevent losing the true focus of this study, we decided to only perform one selection-fine-tuning cycle.

\section{Results}\label{sec:results}

We divide our experiments into two settings: staff-level and full-page. For the full-page scenario, we use the \gls{smt} model on the \Mozarteum{} and \Polonia{} datasets, as these collections have only been addressed using \gls{la}-free pipelines in prior work \citep{smtpp}. The remaining corpora have been studied exclusively with \gls{la}-based approaches, so we use the \gls{crnn} architecture accordingly. Exploring \gls{la}-free models for these datasets is beyond the scope of this study.

This section provides a structured analysis of the experimental results. Section~\ref{sec:global-summary} presents an overall summary, comparing performance across different annotation budgets and highlighting the trade-off between labeling effort and error reduction. Section~\ref{sec:uncertainty-results} examines the similarity between confidence-based and error-based selection strategies, and compares the proposed uncertainty sampling criteria. Finally, Section~\ref{sec:diversity-vs-uncertainty} evaluates diversity-based sampling and contrasts it with uncertainty-based approaches.

Despite defining the annotation budget as a fixed number of samples, it is important to emphasize the significant disparity in annotation effort between the two scenarios. While annotating a single staff-line sample involves, on average, 26 tokens, a full-page sample may contain approximately 2,108 tokens. Therefore, results from these two settings are not directly comparable, and the analyses presented in the following sections focus solely on comparing sampling criteria within each scenario.

\subsection{Overall results}\label{sec:global-summary}

To contextualize the performance of active learning, we define two reference points: (1) the zero-shot setting, which corresponds to the performance of the pre-trained model without any fine-tuning, and (2) the oracle setting, where the model is fine-tuned using all available training samples from the target dataset. These serve as lower and upper bounds on the error rate, respectively. Note that no result is reported for \Mozarteum{} and \Polonia{} under a 100-sample budget, as their official training splits contain fewer than 100 samples.

As shown in Table~\ref{tab:summary-table}, handwritten collections exhibit significantly higher error rates in the zero-shot setting, reflecting the limited generalization of the pre-trained model to historical manuscripts. In contrast, printed collections show similar zero-shot performance in both staff-level and full-page settings. Regarding the oracle baseline, we observe notably higher \gls{ser} values for \Capitan{}, \Mozarteum{}, and \Polonia{}, suggesting that these datasets are inherently more challenging to transcribe.





\begin{table}[hbt]
    \centering
    \renewcommand{\arraystretch}{1}
    \setlength{\tabcolsep}{5pt}
\begin{tabular}{ll c c cccc c}
    \toprule[1pt]
    \multicolumn{2}{l}{\multirow{2}{*}{\textbf{Dataset}}} & \multirow{2}{*}{\textbf{Zero-shot} \phantom{-}} & \multicolumn{4}{c}{\textbf{Budget}} & \multirow{2}{*}{\phantom{-} \textbf{Oracle}\phantom{-}}\\
    \cmidrule{4-7}
     & &  & \textbf{10} & \textbf{25} & \textbf{50} & \textbf{100} & \\
    \cmidrule{1-8}
    \multicolumn{8}{l}{\textbf{\textit{Staff-level}}} \\
    & \Capitan{} & 95.70 & 55.61 & 45.08 & 30.80 & 21.55 & \phantom{0}8.61 \\
    & \Guatemala{} & 84.74 & 36.67 & 24.13 & 19.23 & 14.69 & \phantom{0}2.44 \\
    & \Magnificat{} & 42.20 & 19.81 & 14.75 & \phantom{0}8.73 & \phantom{0}5.65 & \phantom{0}2.95 \\
    & \Mottecta{} & 38.32 & 24.89 & 17.20 & 12.73 & \phantom{0}9.14 & \phantom{0}5.92 \\
    & \Seils{} & 47.41 & 23.41 & 16.38 & \phantom{0}9.71 & \phantom{0}6.70 & \phantom{0}2.85 \\
    & Average & 61.67 & 32.08 & 23.51 & 16.24 & 11.54 & \phantom{0}4.55 \\
    \cmidrule{1-8}
    \multicolumn{8}{l}{\textbf{\textit{Full-page}}} \\
    & \Mozarteum{} & 75.20 & 36.41 & 25.28 & 18.04 & - & 16.36 \\
    & \Polonia{} & 61.61 & 43.33 & 34.42 & 27.24 & - & 24.85 \\
    & Average & 68.40 & 39.87 & 29.85 & 22.64 & - & 20.61 \\
    \bottomrule[1pt]
\end{tabular}
    \caption{\gls{ser} (\%) obtained by the reference settings (zero-shot and oracle), along with the best performance achieved for each annotation budget. Averages for the staff-level and full-page scenarios are also reported.}
    \label{tab:summary-table}
\end{table}


The relative improvement from fine-tuning with low annotation budgets is consistent across collections, regardless of their size. Although the resulting error rate remains high compared to the oracle \gls{ser}, the average error reduction is still remarkable: only 18.24\% of the zero-shot \gls{ser} remains when using 100 samples for staff-level collections and only 34.1\% remains for full-page collections using 50 samples.

\begin{figure}[!htbp]
    \centering
    \includegraphics[width=0.65\linewidth]{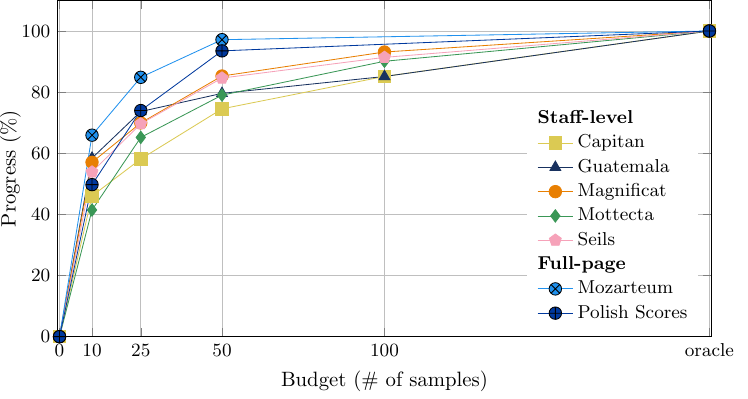}
    \caption{Relative error reduction from zero-shot to oracle performance for each annotation budget.}
    \label{fig:progress}
\end{figure}

To better illustrate the trade-off between annotation cost and model improvement, we reformulate the results from Table~\ref{tab:summary-table} by computing the relative progress toward oracle performance for each budget. This value is normalized such that the zero-shot model corresponds to 0\% progress and the oracle model to 100\%.

Despite the difference in size of the collections, Figure~\ref{fig:progress} shows how the achievable error reduction grows logarithmically with respect to the budget.
This observation has practical implications. If we estimate the total number of corrections required---considering both the annotated symbols used for training and the errors remaining after inference---, it becomes clear that full annotation (oracle setting) is not the most efficient strategy. Maximizing transcription quality with minimal effort requires balancing the cost of manual labeling against the cost of post-processing errors. In this context, the logarithmic trend shown in Figure~\ref{fig:progress} highlights that training with the full dataset yields diminishing returns, making it counterproductive in low-resource scenarios.

\subsection{Comparison of uncertainty sampling criteria}\label{sec:uncertainty-results}

In addition to the main baselines, we include ``oracle sampling'' variants for each uncertainty criterion. These strategies use the true \gls{ser} as a selection signal, following the approach of \citet{active-learning-difficulty-confidence-based}. While not feasible in real-world applications, since they require access to ground-truth labels, these error-based variants serve as upper-bound references for evaluating the effectiveness of practical, confidence-based alternatives.

By expressing both model performance and sampling uncertainty using the same metric, we can assess how closely confidence-based selection approximates the oracle behavior, regardless of how close confidence can estimate the actual error rate. This comparison reveals whether prediction confidence is a reliable proxy for actual error when guiding sample selection. Note that, since we are analyzing these samplings as "uncertainty", high confidence means low uncertainty, whereas high error means high uncertainty. To make the analysis easier to follow, we will use the term "uncertainty" to unify the terminology.

To perform this comparison, we compute the difference in \gls{ser} between both approaches for each criterion (higher values mean confidence worked better). Table~\ref{tab:confidence-vs-error} summarizes these results, averaged across collections. No consistent trend emerges across all criteria, but distinct patterns can be observed.
For instance, confidence-based sampling tends to underperform its error-based counterpart when using low-uncertainty criteria: Lowest, Near $Q_1$, and even Near Mean. Conversely, when sampling based on high uncertainty, confidence-based selection tends to outperform the error-based version.
This asymmetry can be explained by the well-known tendency of neural network models to produce overconfident predictions \citep{neural-network-calibration,focal-loss-calibration,htr-omr-calibration}. As a result, confidence values are less reliable in high-certainty regions, making them a poor proxy for true error when selecting highly confident samples. In contrast, confidence becomes more informative in low-certainty regions, even though the final performances are still very different. These findings are reflected in Table~\ref{tab:confidence-vs-error}, where confidence-based sampling surpasses error-based selection as we move towards higher uncertainty samples. However, these differences seem to become less notable as the annotation budget increases.

\begin{table}[h]
    \centering

\begin{tabular}{w{c}{1.5cm} *{5}{w{c}{1.5cm}}}
    \toprule[1pt]
    \multirow{2}{*}{Budget} & \multicolumn{5}{c}{Uncertainty} \\
    \cmidrule{2-6}
     & Lowest & Near $Q_1$ & Near Mean & Near $Q_3$ & Highest \\
    \cmidrule{1-6}
    10 &   \phantom{-}1.72 & -4.20 & -0.94 & 1.93 &  \phantom{0}8.68 \\
    25 &  -1.46 & -0.39 & -1.60 & 7.01 &  \phantom{0}7.22 \\
    50 &  -2.75 & -2.02 & -2.22 & 3.85 & 10.49 \\
    100 & -1.59 & -1.82 &  \phantom{-}0.36 & 1.86 &  \phantom{0}1.62 \\
    \bottomrule[1pt]
\end{tabular}
    \caption{Difference in \gls{ser} between error-based and confidence-based sampling for each annotation budget. Positive values indicate that the error-based criterion resulted in higher \gls{ser}.}
    \label{tab:confidence-vs-error}
\end{table}

\begin{figure}[!htbp]
    \centering
    \begin{subfigure}{.49\textwidth}
        \includegraphics[width=\textwidth]{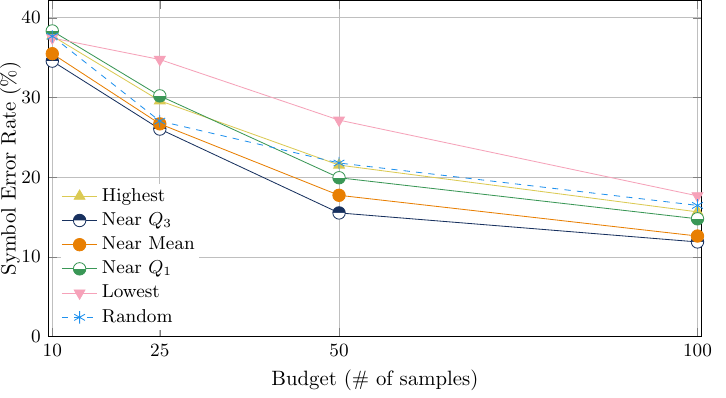}
    \end{subfigure}
    \begin{subfigure}{.49\textwidth}
        \includegraphics[width=\textwidth]{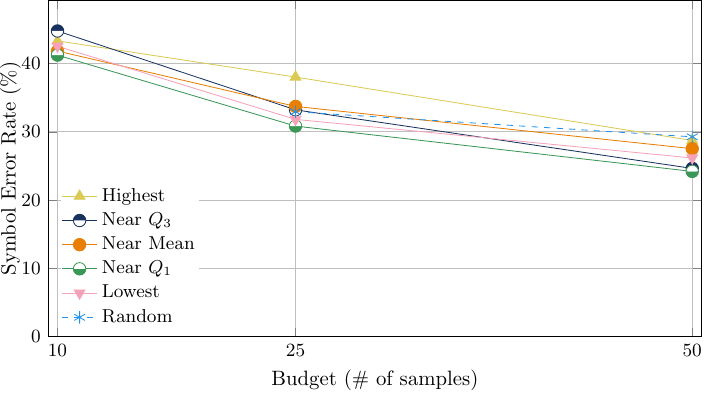}
    \end{subfigure}
    \caption{Average \gls{ser} (\%) obtained using each confidence-based sampling criterion, as well as random sampling, for each annotation budget. Results for \gls{la}-based (left) and \gls{la}-free (right) scenarios are shown separately.}
    \label{fig:uncertainty-plot}
\end{figure}

Although error-based criteria provide a valuable reference, they are not applicable in practical settings. Therefore, the remainder of our analysis on uncertainty sampling focuses exclusively on confidence-based strategies. Figure~\ref{fig:uncertainty-plot} shows the average performance achieved by each uncertainty criterion across the evaluated annotation budgets. 
As a reference, we evaluate a random sampling baseline, in which samples are selected uniformly at random for each labeling budget. To account for its variability, this procedure is repeated 10 times, and we report the average performance.

Most prior work on uncertainty sampling emphasizes selecting the most uncertain samples. However, our results show that this strategy does not consistently yield the best performance. In both \gls{la}-based and \gls{la}-free scenarios, alternative criteria lead to better outcomes. For the former scenario, selecting samples whose confidence lies near the mean or third quartile yields the lowest \gls{ser}. In contrast, the experiments of the second scenario appear to benefit more from samples with relatively low uncertainty. Overall, neither extreme of the uncertainty spectrum demonstrates superior performance or consistency. Moreover, both highest and lowest uncertainty criteria perform similarly to, or even worse than, random sampling.

Based on these observations, we argue that both the highest and lowest uncertainty criteria are likely contaminated by outliers, harming the final performance achievable by training on those samples. Regarding the discrepancy between \gls{la}-based and \gls{la}-free scenarios, it is important to take into account how different these pipelines are, which makes it reasonable for them to benefit from different strategies. \gls{la}-based transcription bounds its input space to single-staves, and performs \gls{ctc}-based decoding. It may have a better generalization potential thanks to \gls{la}, which would explain why it benefits less from samples the model already confidently transcribes. The \gls{la}-free pipeline is based on a transformer decoder, which is known to be very data-hungry, thus making it reasonable for it to benefit from keeping the training set as stable as possible.

While performance differences diminish as the annotation budget increases, the choice of sampling criterion plays a more critical role at lower budgets. In this case, the \textit{near} $\mathit{Q_3}$ (for \gls{la}-based) and \textit{near} $\mathit{Q_1}$ (for \gls{la}-free) strategies consistently achieve superior results across datasets, making them a favorable choice when annotation resources are limited.

\subsection{Comparison of uncertainty sampling and diversity sampling}\label{sec:diversity-vs-uncertainty}

\begin{figure}[!htbp]
    \centering
    \begin{subfigure}{.49\textwidth}
        \includegraphics[width=\textwidth]{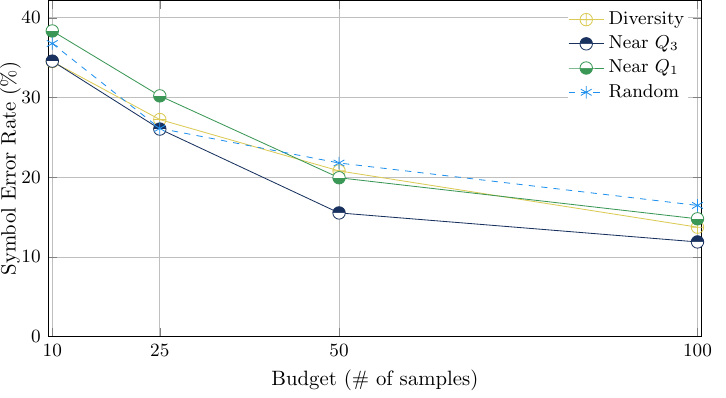}
    \end{subfigure}
    \begin{subfigure}{.49\textwidth}
        \includegraphics[width=\textwidth]{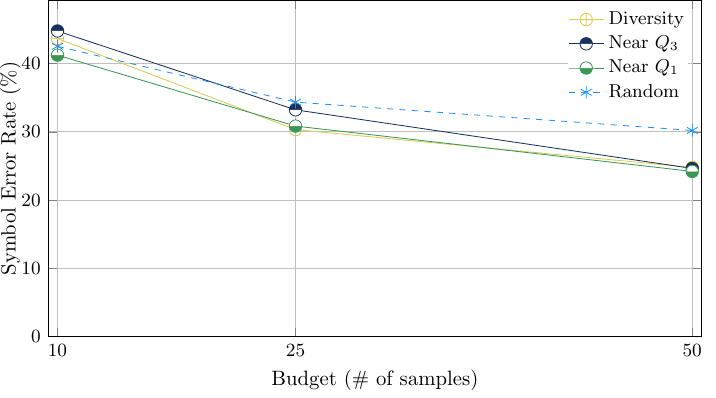}
    \end{subfigure}
    \caption{Average \gls{ser} (\%) obtained using \textit{near}~$\mathit{Q_1}$ and \textit{near}~$\mathit{Q_3}$ uncertainty sampling, diversity sampling, and random sampling for each annotation budget. Results for staff-level (left) and full-page (right) collections are shown separately.}
    \label{fig:diversity-plot}
\end{figure}

Confidence-based uncertainty sampling shows strong potential in fine-tuning setups. However, diversity-based selection offers a compelling alternative. Figure~\ref{fig:diversity-plot} presents the results of diversity sampling, following the same format as in the previous section. For comparison, we include the \textit{near}~$\mathit{Q_1}$ and \textit{near}~$\mathit{Q_3}$ uncertainty criteria as representative baselines, as well as the random sampling reference.

For the \gls{la}-based scenario, diversity sampling is almost tied with random sampling, but fails to match \textit{near}~$\mathit{Q_3}$ consistently as annotation budget grows. However, it aligns more closely with \textit{near}~$\mathit{Q_1}$ as the budget increases, outperforming random sampling, making it a more stable choice than random sampling despite being suboptimal. In contrast, for the \gls{la}-free scenario, diversity sampling starts worse than random sampling, but rapidly matches \textit{near}~$\mathit{Q_1}$, the best-in-slot for this scenario. With this in mind, we would like to highlight that, despite diversity sampling failing to stand out in each scenario, it shows some potential to be more reliable across different scenarios.

Despite its potential, diversity sampling involves significantly higher computational cost, particularly due to feature extraction and clustering. Based on our results, a diversity sampling method with sufficiently solid theoretical backing may achieve performances far more consistent than uncertainty-based sampling. However, such a sampling method must use lightweight calculations to really become a reasonable choice for low-resource environments. For now, however, simpler uncertainty-based approaches remain more efficient and competitive.

\subsection{Discussion}


This study aimed to investigate how different \gls{al} strategies perform in \gls{omr} tasks across a variety of model architectures and data conditions. Beyond just reporting performance results, our findings offer insights into when and why certain AL strategies are more effective, and what limitations must be considered for broader applicability.

Section~\ref{sec:uncertainty-sampling} exposes a potential gap in the literature of uncertainty-based \gls{al}. The commonly used \textit{highest uncertainty} criterion overlaps conceptually with anomaly detection from other fields~\citep{uncertainty-anomaly-detection}, where outliers or noisy samples tend to receive highly uncertain predictions. In our experiments, fine-tuning with samples selected using the \textit{near}~$\mathit{Q_3}$ criterion consistently outperformed the highest uncertainty strategy. This suggests that selecting moderately uncertain samples---rather than the most extreme ones---may help avoid noise and improve learning efficiency, especially in low-budget scenarios.

Complementing uncertainty sampling, our results also show that diversity-based sampling has some potential, but it is still too expensive for the gains it offers. This fact may be attributed to the neural architecture used for each case, both due to feature extraction cost and due to how closely it follows each \textit{best-in-slot}. The criterion used in our experiments was originally developed by~\citet{active-learning-finetuning} for classification tasks, where it operates on classifier logits that summarize the input globally. In our full-page \gls{omr} setup, the encoder-decoder model (\gls{smt}) shares this structural characteristic: the encoder produces a global context vector used by the decoder to generate the output sequence. This architectural parallel likely explains the strong performance of diversity-based sampling in this scenario. Conversely, in the staff-level scenario, the \gls{crnn} model generates localized, column-wise visual features optimized for \gls{ctc}-based alignment. These features, with positional information for alignment, may be less suitable for global similarity comparisons, which may limit the benefits of diversity in that context.

In this paper, we considered a total of seven datasets with diverse characteristics, resulting in a wide range of experiments to support the conclusions drawn. These datasets include both handwritten and typeset music, providing variability in notation style, complexity, and visual features. However, note the vast heterogeneity that exists within the domain of music, particularly in the context of historical works. Variability arises not only from the writing style of the composer or the historical period, but also from the use of different staff systems (e.g., pentagrams vs. tetragrams) and notations \cite{omr-review,chinese-notation,korean-notation}, the presence or absence of lyrics, decorative elements, or other graphical artifacts. This diversity is virtually infinite, and thus cannot be fully captured within the scope of a single study. For this reason, extending the evaluation to additional datasets that reflect a broader range of styles and historical contexts would be a valuable direction for future research. Furthermore, given the large number of experiments, we deliberately limited our analysis to a single round of \gls{al}, following the approach taken in other work in the literature~\citep{active-learning-finetuning}. This decision allowed us to maintain a focused experimental framework and avoid introducing additional variables that could obscure the main contributions of the study. Nonetheless, exploring multi-round \gls{al} strategies remains an important avenue for future investigation.

Finally, while this work focuses on \gls{omr}, the selection strategies investigated do not rely on domain-specific features. This fact opens the possibility of transferring these findings to other sequence-to-sequence recognition tasks. For instance, uncertainty-based sampling could benefit CRNN-based systems for handwritten text or scene text recognition, whereas diversity-based strategies may be more suitable for encoder-decoder or transformer-based models used in mathematical formulations or multi-line transcription tasks. Extending this framework to such domains represents a promising direction for future work.

\section{Conclusions}\label{sec:conclusions}
In this paper, we presented a study on the use of \glsentryfull{al} for sample-efficient fine-tuning of \glsentryfull{omr} systems. Our experiments covered two scenarios: the \glsentryfull{la}-based and the \gls{la}-free. The results provide empirical evidence that model performance increases logarithmically with the number of labeled samples. This suggests that a small number of informative and non-redundant examples is sufficient to achieve substantial gains through fine-tuning.

We compared several confidence-based uncertainty sampling criteria and demonstrated that selecting the most uncertain samples is not always the most effective strategy. In practice, other heuristics, such as selecting samples near the third quartile, often lead to better results. We also found that confidence-based selection performs similarly to error-based (oracle) sampling, reinforcing the notion that model confidence is a reasonable proxy for sample informativeness. Furthermore, we evaluated diversity sampling, concluding that \textit{near}~$\mathit{Q_3}$ performs best in staff-level scenarios (with a \gls{la}-based pipeline), while full-page tasks (with a \gls{la}-free approach) benefit more from \textit{near}~$\mathit{Q_3}$, closely followed by diversity-based approaches.

Our experimental setup included two kinds of music collections: staff-line mensural scores and full-page images in modern notation. All of them support the hypothesis that highest uncertainty is not the most effective sampling criterion for selecting informative examples. Furthermore, all collections benefit from \gls{al} strategies. 

Although our study addresses key questions about the application of active learning to \gls{omr}, it also highlights promising directions for future research. For instance, combining active learning with self-supervised representation learning, as proposed by \citet{self-supervised-active-learning}, remains unexplored in this context. Another avenue is the integration of unsupervised domain adaptation techniques \citep{unsupervised-domain-adaptation} to better leverage unlabeled data, or developing strategies to estimate upper bounds on the annotation budget required for a target performance to further reduce manual labeling costs. In addition, future research should investigate the transferability of these strategies to other domains and also evaluate their effectiveness on other types of musical scores, such as tablatures, which pose unique challenges for automatic transcription.

\printcredits

\section{Acknowledgements}
The research leading to these results received funding from grant FPU24/00347, funded by\\MICIU/AEI/10.13039/501100011033. It was also supported by "Generalitat Valenciana" through the "Subvenciones a grupos de investigación emergentes" programme (CIGE/2024/212).

\bibliographystyle{cas-model2-names} 

\bibliography{cas-refs}



\end{document}